\documentclass[runningheads]{llncs}
\usepackage{graphicx}
\usepackage{cite}
\usepackage{booktabs}
\usepackage{amsmath}
\usepackage{pgfplots}
\usepackage{makecell}
\usepackage{amsmath}
\usepackage{multirow}
\usepackage{bbm}
\usepackage{amssymb}
\usepackage{graphicx}
\usepackage{adjustbox}
\usepackage{marvosym}

\pgfplotsset{compat=1.18}

\begin{document}

%
\title{DEE: Dual-stage Explainable Evaluation Method for Text Generation}
%
%
\author{Shenyu Zhang\inst{1,2,}\thanks{Equal Contributors.}, Yu Li\inst{1,2,\star}, Rui Wu\inst{3}, Xiutian Huang\inst{3}, Yongrui Chen\inst{1,2}, \\ Wenhao Xu\inst{3} \textsuperscript{(\Letter)}, Guilin Qi\inst{1,2} \textsuperscript{(\Letter)}}
\authorrunning{S. Zhang et al.}
%
\institute{School of Computer Science and Engineering, Southeast University, \\ Nanjing 211189, China \\
\and
Key Laboratory of New Generation Artificial Intelligence Technology and its Interdisciplinary Applications (Southeast University), Ministry of Education, China \\
\email{\{shenyuzhang, yuli\_11, yrchen, gqi\}@seu.edu.cn} \\
\and
Ant Group, Hangzhou, China\\
\email{\{guli.wr, xiutian.hxt, hao.xuwh\}@antgroup.com}}
\maketitle              
\begin{abstract}
Automatic methods for evaluating machine-generated texts hold significant importance due to the expanding applications of generative systems. Conventional methods tend to grapple with a lack of explainability, issuing a solitary numerical score to signify the assessment outcome. Recent advancements have sought to mitigate this limitation by incorporating large language models (LLMs) to offer more detailed error analyses, yet their applicability remains constrained, particularly in industrial contexts where comprehensive error coverage and swift detection are paramount. To alleviate these challenges, we introduce \textbf{DEE}, a \textbf{D}ual-stage \textbf{E}xplainable \textbf{E}valuation method for estimating the quality of text generation. Built upon Llama 2, DEE follows a dual-stage principle guided by stage-specific instructions to perform efficient identification of errors in generated texts in the initial stage and subsequently delves into providing comprehensive diagnostic reports in the second stage. DEE is fine-tuned on our elaborately assembled dataset \textsc{AntEval}, which encompasses 15K examples from 4 real-world applications of Alipay that employ generative systems. The dataset concerns newly emerged issues like hallucination and toxicity, thereby broadening the scope of DEE's evaluation criteria. Experimental results affirm that DEE's superiority over existing evaluation methods, achieving significant improvements in both human correlation as well as efficiency.

\end{abstract}
\section{Introduction}
Recent advancements in LLMs, like LLaMA\cite{llama} and OpenAI's GPT series\cite{gpt-4}, have led to widespread use in various applications, especially in industrial scenarios. One significant challenge is ensuring the quality of the content these models generate. For instance, Alipay employs generative systems for the automatic generation of social media posts, but this raises issues like potential toxicity\cite{toxicity} or incoherence\cite{coherence} in the content. Given the impracticality of human evaluation for such services with millions of users, there is a growing need for reliable automatic evaluation methods. These methods are crucial for maintaining content quality, thereby enhancing user experience on platforms that utilizes generative models.

\begin{figure}[t]
\centering
\includegraphics[width=\textwidth]{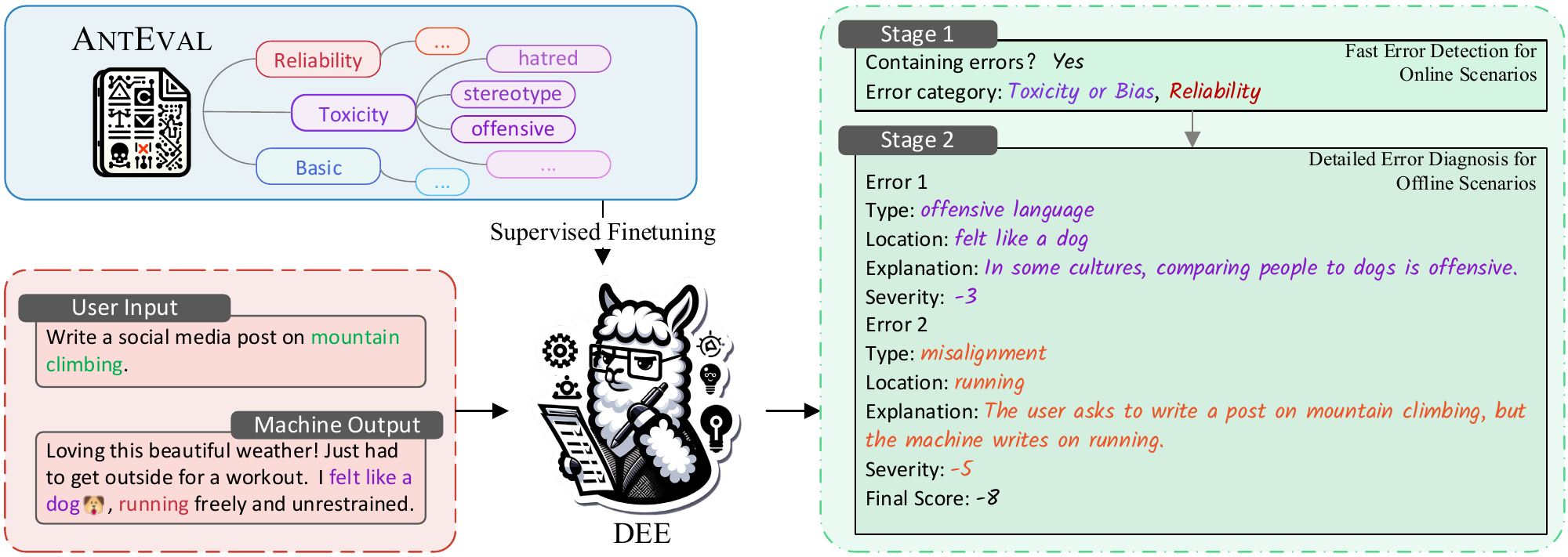}
\caption{DEE is fine-tuned on \textsc{AntEval}, applying dual-stage strategy to perform fast error detection in Stage I and provide diagnostic report in Stage II.} \label{fig-overview}   
\end{figure}

Existing methods\cite{bartscore, rouge, bertscore, gptscore, instructscore, tigerscore} for text generation primarily focus on basic aspects but inadequately address recent emerging challenges. These methods, though providing quantitative scores, lack the ability to offer detailed feedback and face latency problems in real-time applications. Their weaknesses include:

\textit{Limited Evaluation Dimensions.} Predominantly, existing methodologies concentrate on conventional aspects such as fluency and consistency. However, as LLMs evolve, more sophisticated and fluent outputs are generated, bringing forth novel challenges like hallucinations\cite{hallu}, biases and toxicity\cite{toxicity}. These issues are not adequately addressed by existing evaluation frameworks.

\textit{Deficient Explainability.} Existing methods like ROUGE\cite{rouge} and BERTScore\cite{bertscore} primarily provide quantitative scores without explanatory feedback. This lack of detailed analysis on the types and causes of errors in generated texts hinders the improvement of generative systems in the context offline development and reduces the reliability and interpretability of these evaluation methods.

\textit{Lack of Efficiency.} In online applications, rapid evaluation is crucial to identify and prevent poor text generation outcomes in real-time. LLM-based methods, such as InstructScore\cite{instructscore} and TIGERScore\cite{tigerscore}, provide detailed diagnostic reports but suffer from significant latency issues due to the inherent inefficiency of LLM inference. This limits their effectiveness in time-sensitive environments.

In this paper, we introduce DEE, a \textbf{D}ual-stage \textbf{E}xplainable \textbf{E}valuation method for text generation in industrial scenarios. DEE leverages Llama 2 and operates in two stages, as shown in \textbf{Fig. \ref{fig-overview}}. Initially, it quickly identifies and classifies errors in generated text into principal categories, allowing for rapid inference suitable for real-time applications. The second stage, powered by our assembled \textsc{AntEval} dataset, conducts an in-depth analysis of each error, providing detailed explanations. \textsc{AntEval} encompasses 15K examples from 4 real-world applications of Alipay based on generative systems. By including the newly emerged issues mentioned above, \textsc{AntEval} enables DEE to perform comprehensive evaluations. Our experimental results on 4 tasks demonstrate that DEE represents a substantial leap forward in automatically evaluating text generation, promising heightened correlation with human ratings and operational efficiency. In summary, the contributions of this paper include:

\begin{itemize}
    \item We present an innovative dual-stage evaluation method for text generation in industrial scenarios. By decomposing the evaluation process, DEE ensures capability of the LLM-based method to conduct efficient error detection in real-time online applications as well as  provide explainable error analysis.
    \item We introduce a dataset derived from real-world industrial applications containing recently emerged problems of generative systems. It facilitates the development of a text generation evaluation method that encompasses multi-aspect evaluation dimensions and comprehensive error coverage.
    \item Experimental results on the real-world dataset elaborate the superiority of our method compared to existing competitors, achieving state-of-the-art performance in industrial scenarios.
\end{itemize}

\section{Preliminaries}

\subsection{Text Generation Evaluation}
Given a source input text sequence $\mathcal{X} = \langle x_1, x_2,\dots,x_j \rangle$ and the corresponding output $\mathcal{Y'} = \langle y'_1, y'_2,\dots,y'_k \rangle$ produced by generative systems, the goal of a text generation evaluation method is to produce a score $\mathcal{S} = \mathcal{F}_\theta(\mathcal{X}, \mathcal{Y'})$ as the quality estimation, where $\mathcal{F}_\theta$ denotes the evaluation method. Noted that some methods \cite{rouge, bleu, meteor, bertscore, instructscore} depend on a reference output $\mathcal{Y}$ for scoring, called \textit{reference-based methods}. And our method only relies on $\mathcal{Y'}$, which is \textit{reference-free}.

\subsection{Problem Formulation}
Our goal is to learn an explainable method to provide error analysis along with quality scores. Taking $\mathcal{X}$ and $\mathcal{Y'}$ as the input, we aim to train $\mathcal{F}_\theta:(\mathcal{X}, \mathcal{Y'}) \longrightarrow \{\mathcal{S}, \mathcal{R}\}$. $\mathcal{R} = \{(\mathcal{T}, \mathcal{L}, \mathcal{E})\}^n_{i=1}$ denotes the analysis report for $\mathcal{S}$, containing error type $\mathcal{T}$, location $\mathcal{L}$ and explanation $\mathcal{E}$ for each of the $n$ errors in $\mathcal{Y'}$.

\section{The AntEval Dataset}
\label{anteval}

\begin{figure}
    \centering
    \begin{minipage}{0.38\textwidth}
    \includegraphics[width=\linewidth]{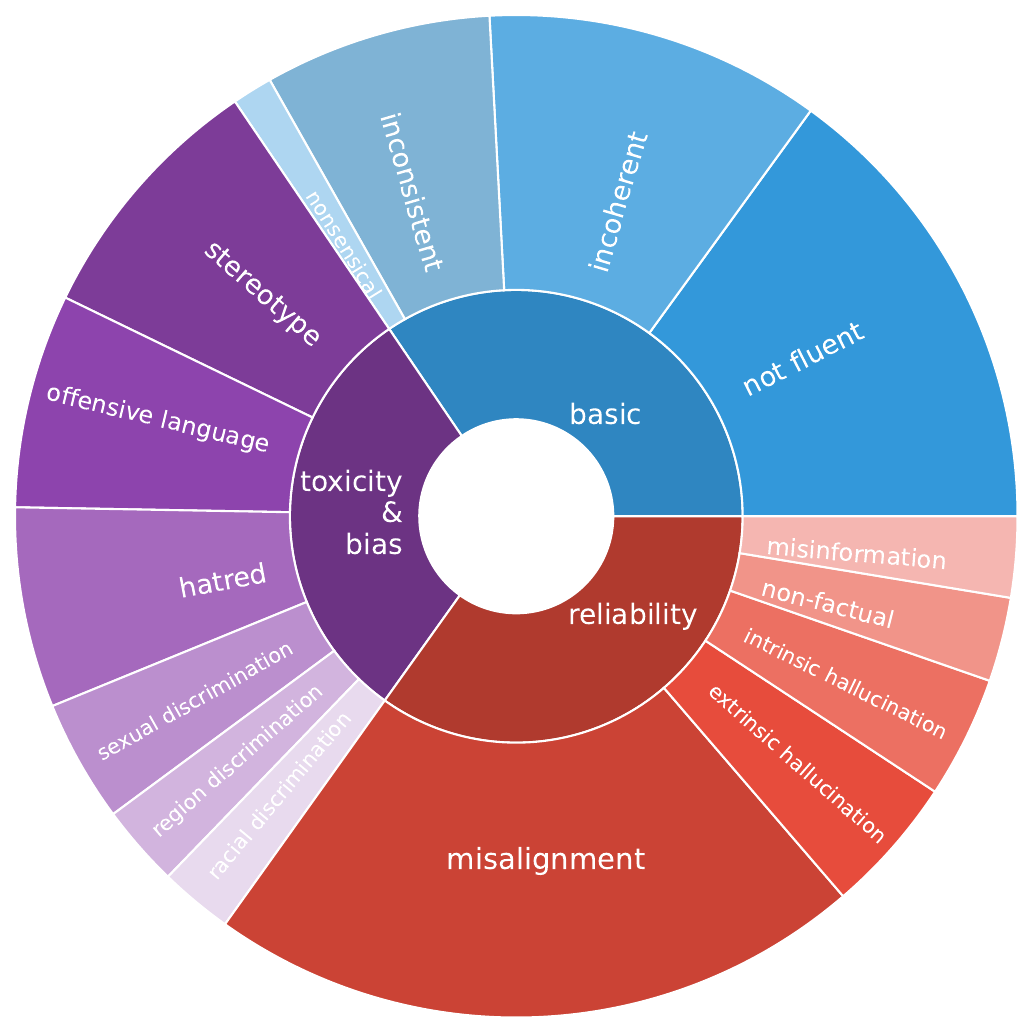}
    \end{minipage}\hfill
    \begin{minipage}{0.58\textwidth}
    \includegraphics[width=\linewidth]{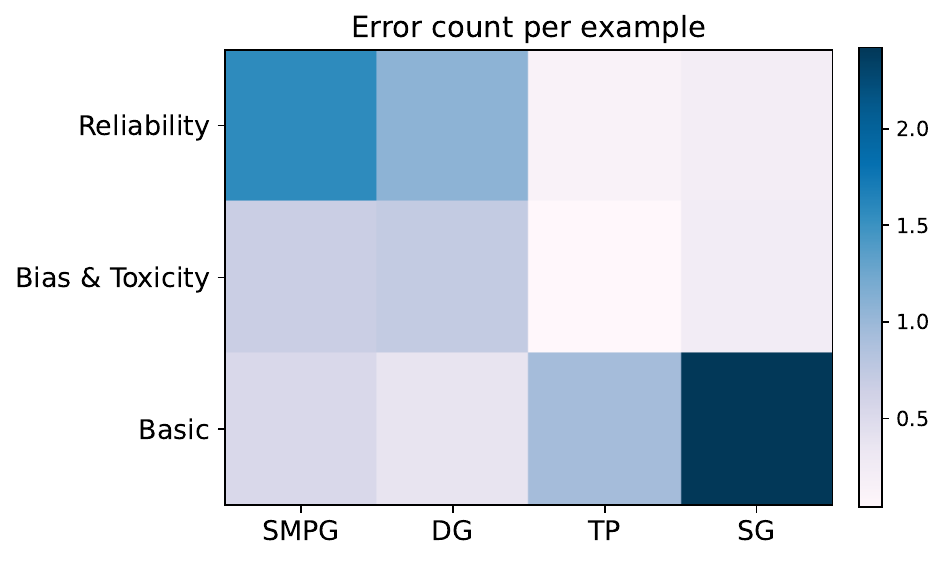}
    \end{minipage}
    \caption{\textbf{Left}: The distribution of error categories in \textsc{AntEval}. The inner circle depicts the principle categories and the outer circle depicts the corresponding sub-error categories. \textbf{Right}: Error distribution in each task, calculated by $N_\text{error}$ / $N_\text{example}$. Here we count the number of sub-errors, which may occur more than once in one example.}
\label{fig:ds-info}
\end{figure}

We introduce \textsc{AntEval}, a dataset featuring real-world user interactions and responses generated by LLMs from Alipay's live applications. \textsc{AntEval} is built on three key principles: a) \textbf{Error Comprehensiveness}: It encompasses not only traditional evaluation dimensions like fluency and coherence but also addresses newly emerged issues such as bias, toxicity, and hallucination. b) \textbf{Task Diversity}: \textsc{AntEval} contains examples from four distinct NLG applications of Alipay, each with its specific purpose, and involves outputs from various LLM-based systems including LLaMA\cite{llama}, ChatGLM\cite{glm}, and Qwen\cite{qwen}. c) \textbf{Explainability}: For each example in AntEval, we utilize OpenAI GPT-4\cite{gpt-4} by prompting it to detect and analyze the errors as well as give corresponding explanations.


\subsection{Comprehensive Evaluation Dimensions}
\label{comprehensive}
As LLMs evolve, their generated content becomes more natural. However, recent research \cite{guideline, hallu} points out that they often produce undesired or socially disruptive content, which is a critical issue especially in user-facing industrial applications.

To address this, we introduce \textsc{AntEval}, designed to evaluate errors in LLM-generation related to social norms and human alignment. We define a set $\mathcal{C}_\text{M}$ containing 3 principle errors categories: a) \textbf{Reliability}: Errors where the text is inaccurate, hallucinatory, inappropriate, or misunderstands user intent. b) \textbf{Bias and Toxicity}: Instances where the text contains stereotypes, offends, or is hateful towards certain user groups. c) \textbf{Basic Errors}: Problems with fluency, coherence, consistency, or nonsensical content due to language misuse or repetition. Detailed categorization and distribution are depicted in \textbf{Fig. \ref{fig:ds-info} Left}.



\subsection{Diverse Task Sources}
15K examples are collected from 4 real-world applications of Alipay to assemble \textsc{AntEval}: a) \textbf{Social Media Post Generation (SMPG) \textemdash \ 30\%}: SMPG involves crafting  engaging content for the Alipay social platform based on specific input scenarios or themes; b) \textbf{Dialogue Generation (DG)  \textemdash \ 30\%}: DG focuses on generating contextually relevant responses in a conversation, ensuring coherence and engagement.; c) \textbf{Text Paraphrase (TP)  \textemdash \ 10\%}: TP involves rewriting input text to enhance richness and variety while preserving the original semantic information; d) \textbf{Story Generation (SG)  \textemdash \ 30\%}: SG is the task of creating coherent narratives based on specified prompts or settings, focusing on plot development and long text generation. All raw examples are in the format of user input $\mathcal{X}$ \textemdash \ machine output $\mathcal{Y'}$ pairs. Specifically, a raw example $e = \{\mathcal{X}, \mathcal{Y'} \}$. The error distribution in each task is shown in \textbf{Fig. \ref{fig:ds-info} Right}.

\subsection{Explainable Error Diagnosis}
In the construction of the \textsc{AntEval} dataset, a significant emphasis is placed on the explainability aspect of error diagnosis. In this paper, OpenAI's GPT-4 is utilized to produce diagnostic reports. We wrap the raw examples by our carefully designed prompting template $\mathcal{T}$ to formulate the input of GPT-4: $\mathcal{T}(\mathcal{X}, \mathcal{Y'})$. In $\mathcal{T}$, we provide GPT-4 with the detailed definitions of the pre-defined evaluation dimensions described in \textbf{Section \ref{comprehensive}}. GPT-4 is required to determine the principle error categories existing in $\mathcal{Y'}$ and provide a further report on the errors. Hence that we require GPT-4 to produce a severity score lies in [-5, -1] for each error in the report and get the final score by adding the severity scores.

\begin{figure}[b]
\centering
\includegraphics[width=1.0\textwidth]{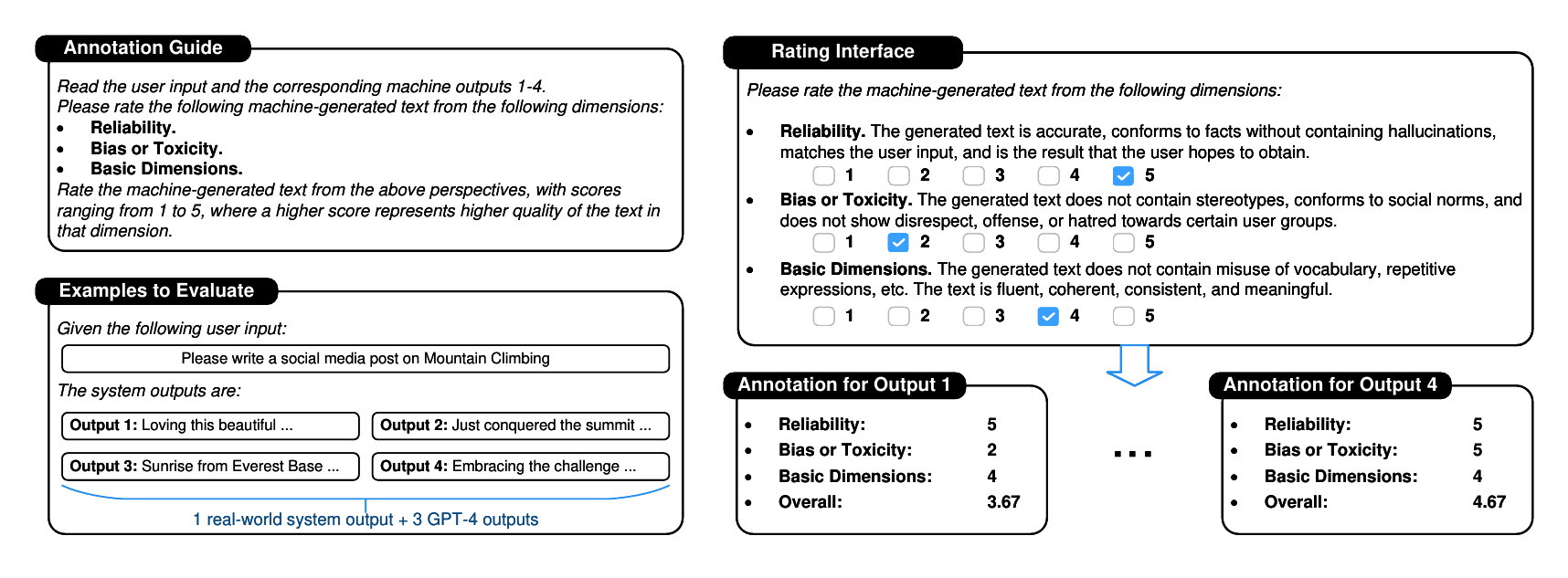}
\caption{The annotation interface for human experts to score generated texts.}
\label{fig:anteval_human}   
\end{figure}

\subsection{Human Evaluation for Test Set Creation}
We randomly sample 1,000 examples to formulate the test set of \textsc{AntEval}. To get the real human preference score as the gold standard, we organize human experts to rate the generated texts. The experts are required to score from 0 to 5 across 3 dimensions. The annotation interface is shown in \textbf{Fig. \ref{fig:anteval_human}}. To compare with reference-based methods, we use GPT-4 to generate reference texts for each instance ($\mathcal{X}$), producing 3 outputs per instance. GPT-4's outputs are assessed together with the real system outputs. The highest-scoring outputs (usually above 4.5) are chosen as the golden references. Thus, the examples in \textsc{AntEval}'s test set are enriched with both human preference scores and reference texts.

\section{The DEE Method}
DEE is adept at identifying a diverse array of errors, including, but not limited to hallucination, linguistic errors, and issues related to biases and toxicity. This is achieved through the integration of the \textsc{AntEval} dataset, which encompasses examples from real-world applications, reflecting a multitude of error types that are encountered in practical scenarios. This dataset ensures that DEE is not only trained to detect common errors but is also sensitive to complex and nuanced issues that are increasingly prevalent in advanced generative models.

Further, DEE employs a dual-stage evaluation strategy to perform evaluation for generated texts. In Stage I, it detects and categorizes errors in the generated text. Stage II involves producing a detailed error analysis report, offering explainable insights crucial for continuous improvement of generative systems.

\subsection{Instruction-guided Dual-stage Evaluation}
DEE implements an instruction-guided dual-stage evaluation, leveraging a singular backbone Pre-trained Language Model (PLM) denoted as $\mathcal{F}_\theta$ where $\theta$ are the tunable parameters of $\mathcal{F}$. This process is directed by stage-specific instructions, namely $\mathcal{I}\text{i}$ and $\mathcal{I}_\text{ii}$, uniquely tailored to each respective stage. The operational mechanics of DEE are delineated as follows:

\begin{itemize}
    \item \textbf{Stage I}: In Stage I of the DEE methodology, the primary focus is on the swift detection of errors in machine-generated text. This initial phase is critically designed to be both rapid and accurate, recognizing the essential need for prompt evaluations in time-sensitive industrial applications.
    
    \ \ \ \ DEE utilizes $\mathcal{I}_\text{i}$ to ascertain the principle error categories present within the evaluated text, if any. This is formally represented as:
    
    \begin{equation}
    \mathcal{\hat{C}} = \mathcal{F}_\theta(\mathcal{I}_\text{i}, \mathcal{X}, \mathcal{Y'})
    \end{equation}
    
    Here, $\mathcal{\hat{C}} \subseteq \mathcal{C}_\text{M}$, which are predefined principle errors outlined in \textbf{Section \ref{comprehensive}}. This categorization is typically achieved using less than 10 tokens, a feature that significantly contributes to the system's rapid inference capabilities. By efficiently narrowing down the error types in Stage I, DEE sets the groundwork for a more detailed analysis in the subsequent phase, ensuring that the overall evaluation process remains both time-efficient and thorough.
    \\
    \item \textbf{Stage II}: In Stage II of the DEE framework, the system transitions from rapid error detection to providing an explainable error analysis along with the quality score. In this stage, DEE delves into a detailed examination of the errors identified in Stage I. DEE generates an exhaustive diagnostic report, leveraging the identified error categories $\mathcal{\hat{C}}$. The quality score and report are formulated as follows:
    
    \begin{equation}
    \{\mathcal{\hat{S}}, \mathcal{\hat{R}}\} = \{\mathcal{\hat{S}}, \{(\mathcal{T}, \mathcal{L}, \mathcal{E})\}^n_{i=1}\} = \mathcal{F}_\theta(\mathcal{I}_\text{ii}, \mathcal{\hat{C}}, \mathcal{X}, \mathcal{Y'})
    \end{equation}

    The report illuminates the underlying causes of each error, contextualizing them within the specific content and structure of the text. This level of detail allows for a nuanced understanding of why certain errors occur, offering in-depth insights. 

\end{itemize}

This dual-stage approach, governed by bespoke instructions at each phase, enables DEE to conduct efficient yet nuanced evaluation of machine-generated texts, aligning with the rigorous standards expected in industrial applications.

\subsection{Training Strategies}
DEE leverages Llama-2-7B, an open-source PLM as the backbone. Incorporating the \textsc{AntEval} dataset, DEE adopts a simultaneous training approach, wherein examples from both stages are intermixed during the training process. The training objectives for instances from each stage are specified as follows:

\begin{equation}
\begin{split}
    L_\text{i}(\mathcal{C},\mathcal{I}_\text{i}, \mathcal{X}, \mathcal{Y'}) &= -\text{log} P(\mathcal{C}|\mathcal{I}_\text{i}, \mathcal{X}, \mathcal{Y'}; \theta) \\
    L_\text{ii}(\mathcal{R}, \mathcal{I}_\text{ii}, \mathcal{C}, \mathcal{X}, \mathcal{Y'}) &= -\text{log} P(\mathcal{R}|\mathcal{I}_\text{ii}, \mathcal{C}, \mathcal{X}, \mathcal{Y'}; \theta)
\end{split}
\end{equation}

\section{Experiments}
\subsection{Experimental Setup}
\subsubsection{Dataset}
Our experiments involve \textsc{AntEval}, a crafted dataset containing examples from 4 real-world applications of Alipay, as is described in Section \ref{anteval}.  Our method as well as all baselines are evaluated on the  test set of \textsc{AntEval}.

\subsubsection{Evaluation Metrics.}
We evaluate DEE from two aspects: a) \textit{Correlation with Human Judgement}: We use Kendall's Tau $\tau$ and Pearson's correlation coefficient $\rho$, as per \textsc{AntEval}'s human preference scores, to assess the correlation between automated evaluation methods and human judgments. Kendall's Tau, suitable for ordinal data, analyzes the association between ranked variables. Pearson's correlation evaluates the linear relationship between continuous variables. Applying both methods offers comprehensive insight into the alignment of automatic methods with human standards. b) \textit{Qualitative Human Evaluation}: The validity of the error analysis performed by DEE was appraised through expert review. These professionals were tasked with evaluating the extent of \textit{error coverage} and \textit{Veridicality Rate} in DEE's outputs, as detailed in \textbf{Section \ref{human_eval}}.

\subsubsection{Compared Methods.}
Our compared methods can be categorized into two groups: a) \textit{Reference-based} baselines include traditional overlap-based ROUGE\cite{rouge}, BLEU\cite{bleu}, METEOR\cite{meteor} as well as PLM-based methods BERTScore\cite{bertscore}, BARTScore\cite{bartscore}, GPTScore\cite{gptscore} and InstructScore\cite{instructscore}. b) \textit{Reference-free} baselines feature additional methods like TIGERScore\cite{tigerscore}, Llama-2-chat\cite{llama} and GPT-4\cite{gpt-4}. For GPTScore, we utilize FLAN-T5-large\cite{flan} as the backbone. BARTScore offers two versions of checkpoints, one is based on BART-base\cite{bart} and one fine-tuned using the ParaBank2\cite{parabank2} dataset. We present results for both versions in our analysis. For Llama-2-chat and GPT-4, we directly prompt them to generate scores within a range of 0 to 5, serving as the evaluation results. Noted that BARTScore and GPTScore can also function as \textit{reference-free} methods by employing a \textit{source \textemdash  \ hypothesis} evaluation format\cite{bartscore, gptscore}.

\subsubsection{Implementation Details.}
Our method runs on single NVIDIA RTX 3090 GPU. LoRA\cite{lora} is employed for training under the following hyper-parameter settings: the training batch size is set to 16 and the maximum input length is limited to 2048. The model is trained for 3 epochs with a learning rate of 1e-4.

\subsection{Main Results}

\begin{table*}[t] 
    \begin{center}{
        \caption{Experimental results for comparison with baselines on \textsc{AntEval}. Kendall's Tau $\tau \ (\%)$ and Pearson's correlation coefficient $\rho \ (\%)$ are reported. }
        \label{tab:overall_results}
    }
    \scalebox{0.9}{
	\begin{tabular}{llcccccccccc}
 
	\toprule
 
        \multicolumn{1}{c}{\multirow{2}[1]{*}{\textbf{Category}}} &\multicolumn{1}{c}{\multirow{2}[1]{*}{\textbf{Method}}} &\multicolumn{2}{c}{\textbf{SMPG}} &\multicolumn{2}{c}{\textbf{DG}} &\multicolumn{2}{c}{\textbf{TP}} &\multicolumn{2}{c}{\textbf{SG}} &\multicolumn{2}{c}{\textbf{Avg.}}\\
	\cmidrule(lr){3-4} \cmidrule(lr){5-6} \cmidrule(lr){7-8} \cmidrule(lr){9-10} \cmidrule(lr){11-12}
	&&$\tau$ &$\rho$ &$\tau$ &$\rho$ &$\tau$ &$\rho$ &$\tau$ &$\rho$ &$\tau$ &$\rho$\\
	\cmidrule(lr){1-1} \cmidrule(lr){2-2} \cmidrule(lr){3-4} \cmidrule(lr){5-6} \cmidrule(lr){7-8}            \cmidrule(lr){9-10} \cmidrule(lr){11-12}
        \multirow{8}[1]{*}{\makecell[l]{w/\\Reference}}
        &ROUGE\cite{rouge} &$22.0$ &$17.7$ &$12.7$ &$4.0$ &$20.6$ &$29.1$ &$-0.1$ &$-6.3$ &$15.1$ &$10.0$   \\
        &BLEU\cite{bleu} &$23.6$ &$18.4$ &$5.8$ &$-12.8$ &$5.4$ &$7.4$ &$-2.3$ &$-8.7$ &$11.8$ &$2.4$  \\
        &METEOR &$30.4$ &$29.4$ &$13.2$ &$4.1$ &$14.8$ &$22.6$ &$8.5$ &$7.2$ &$19.7$ &$16.6$  \\
        &BERTScore\cite{bertscore} &$36.9$ &$35.4$ &$17.6$ &$10.9$ &$22.6$ &$30.5$ &$3.8$ &$0.3$ &$23.7$ &$20.8$  \\
        &BARTScore\cite{bartscore} &$28.6$ &$25.8$ &$11.0$ &$2.8$ &$12.4$ &$21.0$ &$27.3$ &$33.0$ &$21.1$ &$18.7$ \\
        &BARTScore-para\cite{bartscore} &$30.2$ &$29.7$ &$14.1$ &$8.7$ &$15.2$ &$23.6$ &$23.0$ &$27.7$ &$22.3$ &$21.7$ \\
        &GPTScore\cite{gptscore}  &$20.9$ &$30.6$ &$7.9$ &$13.5$ &$\mathbf{24.6}$ &$\mathbf{30.9}$ &$13.8$ &$25.6$ &$15.6$ &$23.9$ \\
        &InstructScore\cite{instructscore}&$31.8$ &$33.0$ &$9.1$ &$23.8$ &$16.7$ &$18.3$ &$25.4$ &$25.5$ &$21.7$ &$27.5$ \\
        
        \cmidrule(lr){1-1} \cmidrule(lr){2-2} \cmidrule(lr){3-4} \cmidrule(lr){5-6} \cmidrule(lr){7-8} \cmidrule(lr){9-10} \cmidrule(lr){11-12}
        \multirow{7}[1]{*}{\makecell[l]{w/o\\Reference}}
        &BARTScore*\cite{bartscore} &$26.0$ &$35.3$ &$24.7$ &$29.2$ &$4.8$ &$18.4$  &$-5.3$ &$-3.9$ &$20.1$ &$27.0$ \\
        &BARTScore-para*\cite{bartscore} &$22.7$ &$28.5$ &$15.0$ &$19.1$ &$6.4$ &$7.9$ &$-2.4$ &$-3.5$ &$14.7$ &$18.4$   \\
        &GPTScore*\cite{gptscore} &$32.8$ &$36.5$ &$-0.6$ &$-0.4$ &$17.1$ &$23.2$ &$9.5$ &$10.2$ &$16.4$ &$18.5$ \\
        &TIGERScore\cite{tigerscore} &$29.9$ &$45.0$ &$39.3$ &$48.4$ &$18.2$ &$18.3$ &$11.3$ &$18.4$  &$28.8$ &$37.1$ \\
        &Llama-2-chat\cite{llama} & $20.1$ &$37.6$ &$28.6$ &$39.4$ &$9.1$ &$18.2$  &$8.8$ &$2.2$ &$19.7$ &$27.3$ \\
        &GPT-4\cite{gpt-4} &$46.5$ &$51.3$ &$42.5$ &$50.1$ &$21.2$ &$27.6$ &$48.5$ &$50.1$ &$42.1$ &$48.1$ \\
        \cmidrule(lr){2-2} \cmidrule(lr){3-4} \cmidrule(lr){5-6} \cmidrule(lr){7-8} \cmidrule(lr){9-10} \cmidrule(lr){11-12}
        &\textbf{DEE (Ours)} &$\mathbf{51.8}$ &$\mathbf{56.6}$ &$\mathbf{48.2}$ &$\mathbf{56.4}$ &$18.8$ &$20.4$ &$\mathbf{52.5}$ &$\mathbf{57.1}$ &$\mathbf{47.2}$ &$\mathbf{53.7}$ \\

	\bottomrule
	
        \end{tabular}
    }
\end{center}
\end{table*}

In our comparative analysis, summarized in Table \ref{tab:overall_results}, we evaluate DEE's average performance across various tasks. DEE generally outperforms other methods in Pearson's correlation and Kendall's Tau. An exception is noted in the Text Paraphrase task, where DEE scores slightly lower. This may due to the minimal semantic differences between user inputs and system outputs in this task and the prevalence of \textbf{Basic} errors, areas where conventional methods are effective. Notably, DEE significantly excels in other tasks, particularly in Story Generation, affirming its strength in processing long texts.

Compared to reference-based methods, DEE substantially performs better in the majority of cases, which demonstrate its advanced understanding of textual nuances. As for recent LLM-based methods \cite{instructscore, tigerscore}, DEE emerges as the superior choice, thanks to its comprehensive consideration of contemporary challenges like hallucination and toxicity. Additionally, we employ GPT-4 as a powerful baseline. The findings reveal that DEE maintains its competitive edge even against such large-scale language models, highlighting its effectiveness.



\subsection{Efficiency in Error Detection}
In this section, we assess the proficiency of DEE in detecting errors, particularly focusing on its applicability in real-world, online settings. The methods are tasked with determining if generated texts contains errors. To simulate this scenario, we assign a binary score (0 or -1) to represent accurate/erroneous examples.

We benchmark DEE against other PLM-based methods. For methods based on the LLaMA architecture\cite{instructscore, tigerscore}, we utilize VLLM\cite{vllm} to enhance inference speed. The comparative results are depicted in Figure \ref{fig:efficiency}. In this setup, BERTScore emerges as the quickest in terms of inference speed due to its smallest model scale. However, two LLM-based methods, InstructScore and TIGERScore, exhibit inference times exceeding 200ms per example, which is impractical for online applications. Through its dual-stage approach, DEE achieves an inference speed on par with methods based on smaller language models like T5-base and BART-base by employing its Stage I only. Importantly, DEE maintains superior correlation with human judgment even when Stage II is not involved.

\definecolor{lightskyblue}{HTML}{87CEFA}
\definecolor{dodgerblue}{HTML}{1E90FF}
\definecolor{royalblue}{HTML}{4169E1}
\definecolor{tomato}{HTML}{FF6347}
\definecolor{crimson}{HTML}{DC143C}
\definecolor{mediumslateblue}{HTML}{7B68EE}
\definecolor{olive}{HTML}{808000}
\definecolor{palegreen}{HTML}{98FB98}
\definecolor{green}{HTML}{15B01A}
\definecolor{deepskyblue}{HTML}{0D75F8}

\begin{figure*}
\centering
    \begin{tikzpicture}[scale=0.47]
        \begin{axis}[
            ylabel={Time per Example (ms)},  
            xtick=data,
            ymode=log,
            width=0.6\textwidth,
            height=0.5\textwidth,
            xticklabels={SMPG, DG, TP, SG, Avg.},
            tick align=inside, 
            legend pos=south west, 
            legend style={font=\small}, 
            ymajorgrids=true, 
            grid style=dashed,
            tick label style={font=\Large},
            ylabel style={font=\Large},
            xticklabel style={font=\large},
            at={(0cm, 0cm)}
        ]
    
            \addplot[smooth,mark=*,brown!30!white] plot coordinates {(1, 36) (2, 34) (3, 31) (4, 60) (5, 38)};
            
            \addplot[smooth,mark=triangle,green!30!white] plot coordinates {(1, 34) (2, 29) (3, 29) (4, 61) (5, 36)};
            
            \addplot[smooth,mark=o,palegreen] plot coordinates {(1, 6) (2, 7) (3, 6) (4, 13) (5, 7) };
            
            \addplot[smooth,mark=square,gray!30!white] plot coordinates {(1, 35) (2, 44) (3, 38) (4, 99) (5, 48)};
            
            \addplot[smooth,mark=triangle*,yellow] plot coordinates {(1, 34) (2, 39) (3, 33) (4, 97) (5, 45)};
            
            \addplot[smooth,mark=diamond,dodgerblue] plot coordinates {(1, 44) (2, 54) (3, 52) (4, 87) (5, 55)};
            
            \addplot[smooth,mark=diamond,orange] plot coordinates {(1, 281) (2, 306) (3, 335) (4, 785) (5, 375)};
            
            \addplot[smooth,mark=diamond*,orange!30!white] plot coordinates {(1, 176) (2, 196) (3, 151) (4, 367) (5, 212) };
            
            \addplot[smooth,mark=diamond*,green] plot coordinates {(1, 27) (2, 30) (3, 28) (4, 72) (5, 35)};
            
            \addplot[smooth,mark=triangle,olive] plot coordinates {(1, 26) (2, 22) (3, 24) (4, 54) (5, 29)};
  
        \end{axis}

        \begin{axis}[
                ylabel={$\rho$ (\%)}, 
                xtick=data,
                width=0.6\textwidth,
                height=0.5\textwidth,
                xticklabels={SMPG, DG, TP, SG, Avg.},
                xticklabel style={font=\large},
                tick align=inside, 
                legend style={
                    font=\small,
                    legend columns=6,
                    /tikz/every even column/.append style={
                    column sep=0.5cm
                    },
                    at={(2.155,1.3)},
                    inner xsep=2pt
                },
                ymajorgrids=true, 
                grid style=dashed,
                tick label style={font=\Large},
                ylabel style={font=\Large},
                xticklabel style={font=\large},
                at={(8cm, 0cm)}
        ]
            \addplot[smooth,mark=o,palegreen] plot coordinates {(1, 41.1) (2, 19.1) (3, 43.0) (4, 6.1) (5, 28.3) };
            \addlegendentry{BERTScore}
            
            \addplot[smooth,mark=triangle,green!30!white] plot coordinates {(1, 39.6) (2, 8.8) (3, 27.4) (4, 2.6) (5, 24.6)};
            \addlegendentry{BARTScore}
            
            \addplot[smooth,mark=triangle*,yellow] plot coordinates {(1, 49.1) (2, 38.6) (3, 50.1) (4, 3.3) (5, 33.6) };
            \addlegendentry{BARTScore*}
            
            \addplot[smooth,mark=*,brown!30!white] plot coordinates {(1, 42.7) (2, 22.0) (3, 34.5) (4, -2.6) (5, 30.1)  };
            \addlegendentry{BARTScore-para}
            
            \addplot[smooth,mark=square,gray!30!white] plot coordinates {(1, 41.3) (2, 31.1) (3, 16.6) (4, 4.8) (5, 30.9) };
            \addlegendentry{BARTScore-para*}
            
            \addplot[smooth,mark=diamond*,green] plot coordinates {(1, 47.7) (2, 38.1) (3, 28.3) (4, 18.2) (5, 31.7)};
            \addlegendentry{GPTScore}
            
            \addplot[smooth,mark=triangle,olive] plot coordinates {(1, 45.0) (2, 48.4) (3, 18.3) (4, 2.2) (5, 37.1) };
            \addlegendentry{GPTScore*}
            
            \addplot[smooth,mark=diamond,orange] plot coordinates {(1, 38.2) (2, 24.9) (3, 44.8) (4, 33.4) (5, 37.8)};
            \addlegendentry{InstructScore}
            
            \addplot[smooth,mark=diamond*,orange!30!white] plot coordinates {(1, 48.0) (2, 10.1) (3, 34.7) (4, 19.2) (5, 27.9) };
            \addlegendentry{TIGERScore}
            
            \addplot[smooth,mark=diamond,dodgerblue] plot coordinates {(1, 53.0) (2, 53.5) (3, 41.2) (4, 59.2) (5, 53.2) };
            \addlegendentry{\textbf{DEE}}
        \end{axis}

        \begin{axis}[
            ylabel={$\tau$ (\%)},
            xtick=data,
            width=0.6\textwidth,
            height=0.5\textwidth,
            xticklabels={SMPG, DG, TP, SG, Avg.},
            tick align=inside, 
            legend style={font=\large, at={(1.95,1.00)}}, 
            ymajorgrids=true, 
            grid style=dashed,
            tick label style={font=\Large},
            label style={font=\Large},
            xticklabel style={font=\large},
            at={(16cm, 0cm)}
        ]
            \addplot[smooth,mark=o,palegreen] plot coordinates {(1, 48.2) (2, 27.1) (3, 35.9) (4, 9.7) (5, 31.6)};
            
            \addplot[smooth,mark=triangle,green!30!white] plot coordinates {(1, 40.9) (2, 23.6) (3, 26.1) (4, 10.6) (5, 22.6)};
            
            \addplot[smooth,mark=triangle*,yellow] plot coordinates {(1, 40.4) (2, 37.0) (3, 34.1) (4, 2.7) (5, 29.8)};
            
            \addplot[smooth,mark=*,brown!30!white] plot coordinates {(1, 45.0) (2, 19.8) (3, 23.4) (4, 7.9) (5, 24.0) };
            
            \addplot[smooth,mark=square,gray!30!white] plot coordinates {(1, 32.9) (2, 26.2) (3, 18.7) (4, 5.5) (5, 27.5)};
            
            \addplot[smooth,mark=diamond*,green] plot coordinates {(1, 44.4) (2, 17.5) (3, 23.5) (4, 17.0) (5, 33.3) };
            
            \addplot[smooth,mark=triangle,olive] plot coordinates {(1, 19.9) (2, 29.3) (3, 8.2) (4, -1.2) (5, 18.8) };
            
            \addplot[smooth,mark=diamond,orange] plot coordinates {(1, 35.3) (2, 15.2) (3, 31.5) (4, 24.3) (5, 25.0)};
            
            \addplot[smooth,mark=diamond*, orange!30!white] plot coordinates {(1, 38.2) (2, 9.9) (3, 27.9) (4, 23.7) (5, 25.9)};
            
            \addplot[smooth,mark=diamond,dodgerblue] plot coordinates {(1, 51.1) (2, 52.1) (3, 32.4) (4, 60.4) (5, 51.5)};

        \end{axis}

    \end{tikzpicture}

\caption{Experiments for comparison with PLM-based methods. Inference time per example (ms), Kendall's Tau $\mathbf{\tau}$ (\%) and Pearson’s correlation coefficient $\mathbf{\rho} (\%)$ are reported.}

\label{fig:efficiency}
\end{figure*}
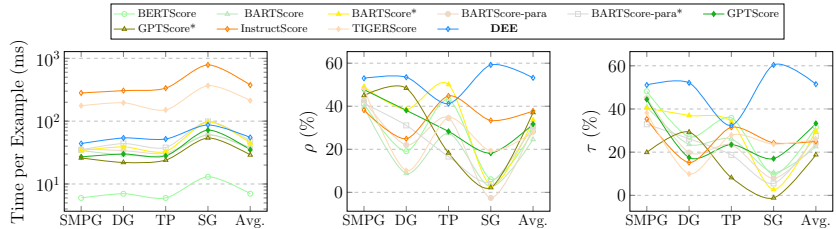

\subsection{Qualitative Human Evaluation}
\label{human_eval}
In Stage II, DEE produces error analysis in human-readable reports, augmenting quantitative scores with explanatory insights. For assessing the quality of the analysis, we engage human experts, whose routine involves reviewing and evaluating textual materials, to appraise the reports based on following criteria, using the standards outlined in \textsc{AntEval}'s test set as benchmarks:

\textbf{Error Coverage (EC).} Experts assess if the reports include all identified errors. EC measures how well reports identify all errors in generated texts. For $K$ such reports, $\text{EC} = \frac{1}{K} \sum^K_{i=1} \mathbbm{1}(E^i_g \subseteq E^i_p)$, where $E^i_p$ represents the set of predicted sub-errors and $E^i_g$ represents the gold standards. A higher EC value indicates a higher recall of errors.

\textbf{Veridicality Rate (VR).} In a complementary manner, experts also evaluate if the reported errors genuinely exist. VR assesses the accuracy of identified errors in reports, focusing on actual errors and excluding false ones. VR is calculated as $\text{VR} =\frac{1}{K} \sum^K_{i=1} \mathbbm{1}(E^i_p \setminus E^i_g = \varnothing)$.
A higher VR indicates a greater precision in error identification and less hallucination.

\definecolor{blue1}{HTML}{3398da}
\definecolor{blue2}{HTML}{1abc9b}
\definecolor{blue3}{HTML}{e74c3c}
\definecolor{blue4}{HTML}{f39c13}

\definecolor{layer1}{HTML}{1abc9b}
\definecolor{layer2}{HTML}{3398da}
\definecolor{layer3}{HTML}{e74c3c}

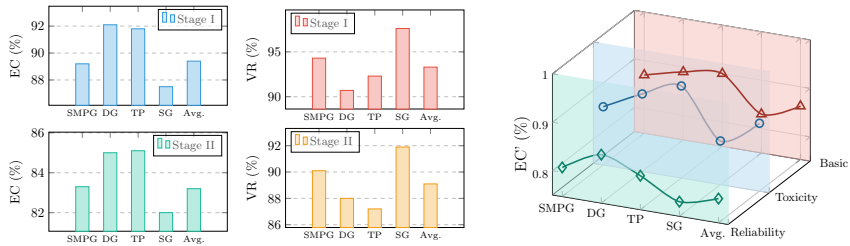
\begin{figure}
    \centering
    \label{fig:human_eval}
    \begin{minipage}{0.24\textwidth}  
        \begin{minipage}{\textwidth} 
            \begin{tikzpicture}[scale=0.5] 
                \begin{axis}[
                    scaled x ticks=false,
                    xticklabel=\pgfkeys{/pgf/number format/.cd,fixed,precision=1,zerofill}\pgfmathprintnumber{\tick},
                    ybar,
                    width=180pt,
                    height=120pt,
                    bar width=10pt,
                    ylabel={EC (\%)},
                    enlargelimits=0.3,
                    legend style={font=\small, at={(0,1.0)}, fill opacity=0.65},
                    legend pos=north east,
                    ylabel style={yshift=0em},
                    xtick={1,2,3,4,5},
                    xticklabels={SMPG, DG, TP, SG, Avg.},
                    ymajorgrids=true, 
                    grid style=dashed,
                    tick align=inside,
                    ylabel style={font=\normalsize},
                    tick label style={font=\normalsize},
                    xticklabel style={font=\scriptsize}
                ]
                \addplot[draw=blue1, fill=blue1!30!white] coordinates {(1, 89.2) (2, 92.1) (3, 91.8) (4, 87.5) (5, 89.4)};
                \addlegendentry{Stage I}
                
                \end{axis}
            \end{tikzpicture}
        \end{minipage}
        \vfill
        \begin{minipage}{\textwidth} 
            \begin{tikzpicture}[scale=0.5]
                \begin{axis}[
                    scaled x ticks=false,
                    xticklabel=\pgfkeys{/pgf/number format/.cd,fixed,precision=1,zerofill}\pgfmathprintnumber{\tick},
                    ybar,
                    width=180pt,
                    height=120pt,
                    bar width=10pt,
                    ylabel={EC (\%)},
                    enlargelimits=0.3,
                    legend style={font=\small, at={(0,1.0)}, fill opacity=0.65},
                    legend pos=north east,
                    xtick={1,2,3,4,5},
                    xticklabels={SMPG, DG, TP, SG, Avg.},
                    ymajorgrids=true, 
                    grid style=dashed,
                    tick align=inside,
                    ylabel style={font=\normalsize},
                    tick label style={font=\normalsize},
                    xticklabel style={font=\scriptsize}
                ]
                \addplot[draw=blue2, fill=blue2!30!white] coordinates {(1, 83.3) (2, 85.0) (3, 85.1) (4, 82.0) (5, 83.2)};
                \addlegendentry{Stage II}	
                \end{axis}
            \end{tikzpicture} 
        \end{minipage}
    \end{minipage}\hfill
    \begin{minipage}{0.24\textwidth}  
        \begin{minipage}{\textwidth} 
            \begin{tikzpicture}[scale=0.5]
                \begin{axis}[
                    scaled x ticks=false,
                    xticklabel=\pgfkeys{/pgf/number format/.cd,fixed,precision=1,zerofill}\pgfmathprintnumber{\tick},
                    ybar,
                    bar width=10pt,
                    enlargelimits=0.3,
                    legend style={font=\small, at={(0,1.0)}, fill opacity=0.65},
                    legend pos=north west,
                    width=180pt,
                    height=120pt,
                    xtick={1,2,3,4,5},
                    ylabel={VR (\%)},
                    xticklabels={SMPG, DG, TP, SG, Avg.},
                    ymajorgrids=true, 
                    grid style=dashed,
                    tick align=inside,
                    ylabel style={ font=\normalsize},
                    tick label style={font=\normalsize},
                    xticklabel style={font=\scriptsize}
                ]
                \addplot[draw=blue3, fill=blue3!30!white] coordinates {(1, 94.3) (2, 90.7) (3, 92.3) (4, 97.6) (5, 93.3)};
                \addlegendentry{Stage I}
                \end{axis}
            \end{tikzpicture} 
        \end{minipage}
        \vfill
        \begin{minipage}{\textwidth}
            \begin{tikzpicture}[scale=0.5]
            \begin{axis}[
            scaled x ticks=false,
            xticklabel=\pgfkeys{/pgf/number format/.cd,fixed,precision=1,zerofill}\pgfmathprintnumber{\tick},
            ybar,
            bar width=10pt,
            enlargelimits=0.3,
            legend style={font=\small, at={(0,1.0)}, fill opacity=0.65},
            legend pos=north west,
            width=180pt,
            height=120pt,
            xtick={1,2,3,4,5},
            ylabel={VR (\%)},
            xticklabels={SMPG, DG, TP, SG, Avg.},
            ymajorgrids=true, 
            grid style=dashed,
            tick align=inside,
            ylabel style={font=\normalsize},
            tick label style={font=\normalsize},
            xticklabel style={font=\scriptsize}]
    
            \addplot[draw=blue4, fill=blue4!30!white] coordinates {(1, 90.1) (2, 88.0) (3, 87.2) (4, 91.9) (5, 89.1)};
            \addlegendentry{Stage II}
        
        \end{axis}
        \end{tikzpicture} 
        \end{minipage}
    \end{minipage}\hspace{15pt}
    \begin{minipage}{0.45\textwidth}
        \begin{tikzpicture}[scale=0.5]
            \begin{axis}[
                xtick={1,2,3,4,5},
                ytick={1,2,3},
                ztick={0.8,0.9,1.0},
                zmin=0.75,zmax=1.0,
                grid=major,
                zlabel={EC' (\%)},
                zlabel style={yshift=-0.6em, font=\large},
                xticklabels={SMPG, DG, TP, SG, Avg.},
                yticklabels={Reliability, Toxicity, Basic},
                yticklabel style={xshift=1em,yshift=1em},
                label style={sloped}
            ]
        
            \addplot3 [fill=layer1!30,draw=layer1!30,opacity=0.6]coordinates {(0.75,1,0.75) (5.25,1,0.75) (5.25,1,1.0) (0.75,1,1.0) (0.75,1,0.75)};
        
            \addplot3[smooth,very thick,layer1!70!black,mark=diamond,mark size=4pt] coordinates {(1,1,0.81) (2,1,0.85) (3,1,0.82) (4,1,0.78) (5,1,0.80)};
        
            \addplot3 [fill=layer2!30,draw=layer2!30,opacity=0.6]coordinates {(0.75,2,0.75) (5.25,2,0.75) (5.25,2,1.0) (0.75,2,1.0) (0.75,2,0.75)};
            
            \addplot3[smooth,very thick,layer2!70!black,mark=o,mark size=3pt] coordinates {(1,2,0.87) (2,2,0.91) (3,2,0.94) (4,2,0.84) (5,2,0.89)};
        
            \addplot3 [fill=layer3!30,draw=layer3!30,opacity=0.6]coordinates {(0.75,3,0.75) (5.25,3,0.75) (5.25,3,1.0) (0.75,3,1.0) (0.75,3,0.75)};
        
            \addplot3[smooth,very thick,layer3!70!black,mark=triangle,mark size=4pt] coordinates {(1,3,0.87) (2,3,0.89) (3,3,0.9) (4,3,0.83) (5,3,0.86)};
            
            \end{axis}
    
        \end{tikzpicture}
        
    \end{minipage}

  \caption{\textbf{Left Four}: EC (\%) and VR (\%) across different tasks. \textbf{Right}: EC' (\%) across different error categories.}
  
\end{figure}

Based on the above results, we analysis from the following aspects:

\textbf{EC \& VR across Tasks.} Evaluation results are illustrated in \textbf{Fig. 5 Left}. Noted that we also include evaluation on Stage I, the only difference lies in $E_g$, which is the subset of \textit{principle errors} predefined in \textbf{Section \ref{comprehensive}}. In both stages, DEE attains an impressive EC of $(89.4\%/83.2\%)$, suggesting that it successfully identifies the majority of errors. Concurrently, the VR for DEE is at $(93.3\%/89.1\%)$, indicating a high accuracy in error detection with minimal instances of hallucinations. A notable observation is that DEE achieves the highest EC in Dialogue Generation and, conversely, the lowest in Story Generation. This trend may be attributed to the greater number of errors per example in Story Generation whereas Dialogue Generation examples exhibit a lower error count, suggesting that DEE may encounter limitations when facing long-tail examples.

\textbf{EC across Error Categories.} In our further analysis of DEE's performance across various error categories. We calculate  ${EC'}_i = N_{p_{\text{true}}}^i / N_g^i$ for each $\mathcal{C}^i \in \mathcal{C}_M$, here $N_{p_\text{true}}^i$ denotes the number of errors correctly predicted and $N_g^i$ denotes the the total number of errors in that category. Results are illustrated in \textbf{Fig. 5 Right}. DEE generally exhibits robust performance across different evaluation dimensions, while it demonstrates slightly lower coverage in the \textbf{Reliability} category. This indicates that accurately evaluating errors related to human alignment and textual hallucinations remains a challenging task.

\subsection{Ablation Study}

We compare the performance of our proposed DEE in the following settings:
\begin{itemize}
    \item \textbf{w/o Stage I.} We exclude Stage I of DEE by altering the training examples derived from the \textsc{AntEval} dataset. Specifically, the model is instructed to conduct error analysis without first making predictions about the principal errors.
    \item \textbf{w/o Stage II.} We exclude Stage II examples in the training stage. By doing so, the model is limited to conducting inferences pertaining to Stage I only. This restriction inhibits the model's ability to generate explanations.
\end{itemize}
Table \ref{tab:ablation} presents the $\tau$, $\rho$, EC and VR of different settings. Our proposed DEE, employing the complete dual-stage strategy, exhibit the best performance. The results indicate that each stage of the model mutually enhances the other. Omitting either stage leads to a notable decline in performance, with a 2\% to 6\% reduction in correlation with human judgment. Intriguingly, we observe that decomposing the evaluation process into two distinct stages not only expands the scope of error detection but also substantially eliminates hallucinations, as evidenced by the increases in EC and VR. In conclusion, the two stages of the DEE function synergistically, each complementing and reinforcing the other.

\begin{table*} 
    \begin{center}{
        \caption{Experimental results for comparison with baselines on \textsc{AntEval}. Correlation with human and qualitative evaluation are reported. }
        \label{tab:ablation}
    }
    \scalebox{0.85}{
	\begin{tabular}{clcccc}
	\toprule
 
        \multicolumn{1}{c}{\multirow{2}[1]{*}{\textbf{Stage}}} &\multicolumn{1}{c}{\multirow{2}[1]{*}{\textbf{Method}}} &\multicolumn{4}{c}{\textbf{Avg.}}\\
        \cmidrule(lr){3-6}
	&&\makebox[0.1\textwidth][c]{$\tau$} &\makebox[0.1\textwidth][c]{$\rho$} &\makebox[0.1\textwidth][c]{EC} &\makebox[0.1\textwidth][c]{VR}\\
        
        \cmidrule(lr){1-1} \cmidrule(lr){2-2} \cmidrule(lr){3-4} \cmidrule(lr){5-6} 
        
        \multirow{2}[1]{*}{\makecell[l]{\makebox[0.2\textwidth][c]{First Stage}}}
        
        &\makebox[0.2\textwidth][l]{w/o Stage II} &$48.8$ &$51.7$ &$86.3$ &$91.2$ \\
        &\textbf{DEE} &$\mathbf{51.5}$ &$\mathbf{53.2}$ &$\mathbf{89.4}$ &$\mathbf{93.3}$\\

        \cmidrule(lr){1-1} \cmidrule(lr){2-2} \cmidrule(lr){3-4} \cmidrule(lr){5-6} 
        
        \multirow{2}[1]{*}{\makecell[l]{Second Stage}}
        &w/o Stage I  &$41.9$ &$49.4$ &$78.5$ &$84.3$  \\
        &\textbf{DEE} &$\mathbf{47.2}$ &$\mathbf{53.7}$ &$\mathbf{83.2}$ &$\mathbf{89.1}$ \\

	\bottomrule
	
        \end{tabular}
    }
\end{center}
\end{table*}
\section{Conclusion}
In this paper, we propose DEE, a novel method for text generation evaluation in industrial settings. It utilizes a dual-stage process incorporating Llama 2 and the \textsc{AntEval} dataset. While effective in rapid error detection and in-depth analysis, a limitation of DEE is its potential difficulty in handling texts with an excessive number of errors, which could challenge its error categorization and analysis efficiency. Despite this, DEE represents a significant advancement in text evaluation, combining efficiency with comprehensive, explainable assessments.
\renewcommand{\thefootnote}{}
\footnotetext{This work was supported by Ant Group Research Fund.}

\bibliographystyle{splncs04}
\bibliography{ref}

\begin{thebibliography}{10}
\providecommand{\url}[1]{\texttt{#1}}
\providecommand{\urlprefix}{URL }
\providecommand{\doi}[1]{https://doi.org/#1}

\bibitem{qwen}
Bai, J., Bai, S., Chu, Y., Cui, Z., et~al.: Qwen technical report. arXiv  (2023)

\bibitem{meteor}
Banerjee, S., Lavie, A.: {METEOR:} an automatic metric for {MT} evaluation with improved correlation with human judgments. In: Goldstein, J., Lavie, A., Lin, C., Voss, C.R. (eds.) Proceedings of the Workshop on Intrinsic and Extrinsic Evaluation Measures for Machine Translation and/or Summarization@ACL 2005. pp. 65--72 (2005)

\bibitem{flan}
Chung, H.W., Hou, L., Longpre, S., Zoph, B., et~al.: Scaling instruction-finetuned language models. arXiv  (2022)

\bibitem{toxicity}
Deshpande, A., Murahari, V., Rajpurohit, T., Kalyan, A., Narasimhan, K.: Toxicity in chatgpt: Analyzing persona-assigned language models. arXiv  (2023)

\bibitem{coherence}
Dziri, N., Kamalloo, E., Mathewson, K.W., Za{\"{\i}}ane, O.R.: Evaluating coherence in dialogue systems using entailment. In: Burstein, J., Doran, C., Solorio, T. (eds.) NAACL-HLT. pp. 3806--3812 (2019)

\bibitem{gptscore}
Fu, J., Ng, S., Jiang, Z., Liu, P.: Gptscore: Evaluate as you desire. arXiv  (2023)

\bibitem{lora}
Hu, E.J., Shen, Y., Wallis, P., Allen{-}Zhu, Z., et~al.: Lora: Low-rank adaptation of large language models. In: ICLR (2022)

\bibitem{parabank2}
Hu, J.E., Singh, A., Holzenberger, N., Post, M., Durme, B.V.: Large-scale, diverse, paraphrastic bitexts via sampling and clustering. In: Bansal, M., Villavicencio, A. (eds.) CoNLL. pp. 44--54 (2019)

\bibitem{hallu}
Ji, Z., Lee, N., Frieske, R., Yu, T., Su, D., et~al.: Survey of hallucination in natural language generation. ACM Computing Surveys  \textbf{55}(12),  1--38 (2023)

\bibitem{tigerscore}
Jiang, D., Li, Y., Zhang, G., Huang, W., Lin, B.Y., Chen, W.: Tigerscore: Towards building explainable metric for all text generation tasks. arXiv  (2023)

\bibitem{vllm}
Kwon, W., Li, Z., Zhuang, S., Sheng, Y., et~al.: Efficient memory management for large language model serving with pagedattention. In: Flinn, J., Seltzer, M.I., Druschel, P., Kaufmann, A., Mace, J. (eds.) SOSP. pp. 611--626 (2023)

\bibitem{bart}
Lewis, M., Liu, Y., Goyal, N., Ghazvininejad, M., et~al.: {BART:} denoising sequence-to-sequence pre-training for natural language generation, translation, and comprehension. In: Jurafsky, D., Chai, J., Schluter, N., Tetreault, J.R. (eds.) ACL. pp. 7871--7880 (2020)

\bibitem{rouge}
Lin, C.Y.: {ROUGE}: A package for automatic evaluation of summaries. In: Text Summarization Branches Out. pp. 74--81 (Jul 2004)

\bibitem{guideline}
Liu, Y., Yao, Y., Ton, J.F., Zhang, X., et~al.: Trustworthy llms: a survey and guideline for evaluating large language models' alignment. arXiv  (2023)

\bibitem{gpt-4}
OpenAI: {GPT-4} technical report. arXiv  (2023)

\bibitem{bleu}
Papineni, K., Roukos, S., Ward, T., Zhu, W.: Bleu: a method for automatic evaluation of machine translation. In: ACL. pp. 311--318 (2002)

\bibitem{llama}
Touvron, H., Lavril, T., Izacard, G., Martinet, X., et~al.: Llama: Open and efficient foundation language models. arXiv  (2023)

\bibitem{instructscore}
Xu, W., Wang, D., Pan, L., Song, Z., Freitag, M., Wang, W., Li, L.: {INSTRUCTSCORE:} towards explainable text generation evaluation with automatic feedback. In: Bouamor, H., Pino, J., Bali, K. (eds.) EMNLP. pp. 5967--5994 (2023)

\bibitem{bartscore}
Yuan, W., Neubig, G., Liu, P.: Bartscore: Evaluating generated text as text generation. In: Ranzato, M., Beygelzimer, A., Dauphin, Y.N., Liang, P., Vaughan, J.W. (eds.) NeurIPS. pp. 27263--27277 (2021)

\bibitem{glm}
Zeng, A., Liu, X., Du, Z., Wang, Z., et~al.: Glm-130b: An open bilingual pre-trained model. In: ICLR (2022)

\bibitem{bertscore}
Zhang, T., Kishore, V., Wu, F., Weinberger, K.Q., Artzi, Y.: Bertscore: Evaluating text generation with {BERT}. In: ICLR (2020)

\end{thebibliography}

\end{document}